\documentclass[runningheads]{llncs}
\usepackage{graphicx}
\usepackage{comment}
\usepackage{amsmath,amssymb} 
\usepackage{color}
\usepackage{hyperref}

\usepackage{tikz}
\usepackage{textcomp}
\usepackage{epsfig}
\usepackage{graphicx,adjustbox}
\usepackage{enumitem}
\usepackage{amsmath}
\usepackage{amssymb}
\usepackage{multirow}
\usepackage{booktabs}
\usepackage{floatrow}
\usepackage{makecell}
\usepackage{pgfplots}
\usepackage[small]{caption}
\usepackage[htt]{hyphenat}
\usepackage{wrapfig}
\usepackage{sidecap}
\DeclareMathOperator*{\argmax}{arg\,max}

\newcommand{\cfbox}[2]{%
    \colorlet{currentcolor}{.}%
    {\color{#1}%
    \fbox{\color{currentcolor}#2}}%
}

\newfloatcommand{capttabbox}{table}[][\FBwidth]

\begin{document}

\pagestyle{headings}
\mainmatter
\def\ECCVSubNumber{6754}  

\title{Learning to Learn Words from Visual Scenes}



\author{D\'idac Sur\'is\inst{1}\thanks{Equal contribution} \and
Dave Epstein\inst{1}$^*$ \and
Heng Ji\inst{2} \and Shih-Fu Chang\inst{1} \and Carl Vondrick\inst{1}}

\authorrunning{D. Sur\'is, D. Epstein et al.}  
%
\institute{\textsuperscript{1}Columbia University \qquad \textsuperscript{2}UIUC\\
\href{http://expert.cs.columbia.edu}{expert.cs.columbia.edu}}

\maketitle

\sidecaptionvpos{table}{b}
\sidecaptionvpos{figure}{c}

\definecolor{new_pink}{RGB}{244,67,54}
\definecolor{new_green}{RGB}{255,235,59}
\definecolor{dark_color}{RGB}{255,0,0}
\definecolor{light_color}{RGB}{167,167,167}

\begin{abstract}

Language acquisition is the process of learning words from the surrounding scene. We introduce a meta-learning framework that \emph{learns how to learn} word representations from unconstrained scenes. We leverage the natural compositional structure of language to create training episodes that cause a meta-learner to learn strong policies for language acquisition. Experiments on two datasets show that our approach is able to more rapidly acquire novel words as well as more robustly generalize to unseen compositions, significantly outperforming established baselines. A key advantage of our approach is that it is data efficient, allowing representations to be learned from scratch without language pre-training. Visualizations and analysis suggest visual information helps our approach learn a rich cross-modal representation from minimal examples.

\end{abstract}

\section{Introduction}

\begin{wrapfigure}[16]{r}{.5\textwidth}

\includegraphics[width=1\linewidth]{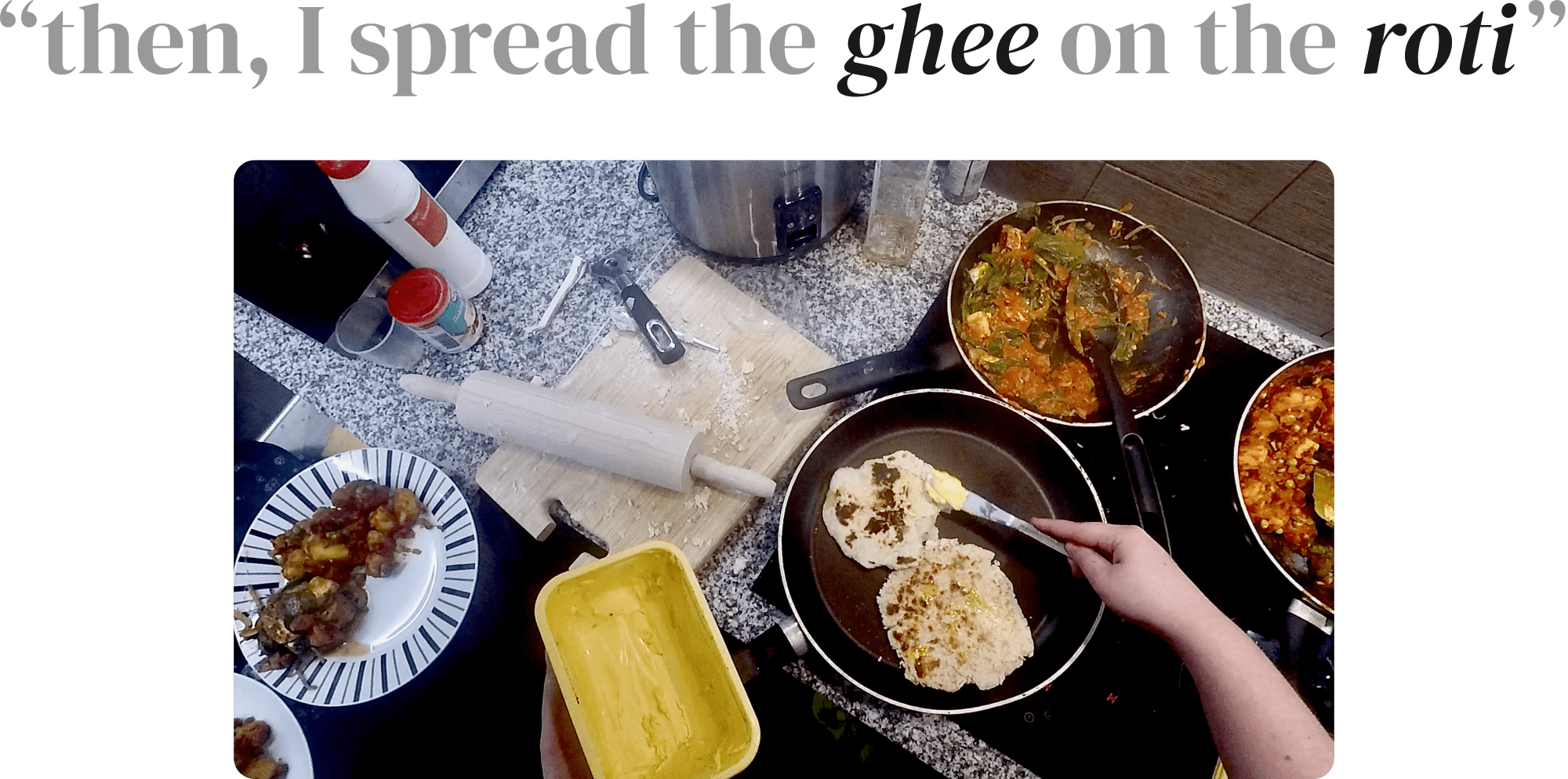}

\caption{\textbf{What is ``ghee'' and ``roti''?} The answer is in the footnote.\protect\footnotemark Although the words ``ghee'' and ``roti'' may be unfamiliar, you are able to leverage the structure of the visual world and knowledge of other words to acquire their meaning. In this paper, we propose a model that learns how to learn words from visual context.}
\label{fig:teaser} 
\end{wrapfigure}
\footnotetext{Answer: \rotatebox[origin=c]{180}{``ghee'' is the butter on the knife, and ``roti'' is the bread in the pan}}

Language acquisition is the process of learning words from the surrounding environment. Although the sentence in Figure 1 contains new words, we are able to leverage the visual scene to accurately acquire their meaning.  While this process comes naturally to children as young as six months old \cite{tincoff1999some} and represents a major milestone in their development, creating a machine with the same malleability has remained challenging. 

The standard approach in vision and language aims to learn a common embedding space  \cite{farhadi2010every,Sun2019a,Li2019b},
however this approach has a number of key limitations. Firstly, these models are inefficient because they often require millions of examples to learn. Secondly, they consistently
generalize poorly to the natural compositional structure of language \cite{gandhi2019mutual}. Thirdly, fixed embeddings are unable to adapt to novel words at inference time, such as in  realistic scenes that are naturally open world. We believe these limitations stem fundamentally from the process that models use to acquire words.

While most approaches learn the word embeddings, we propose to instead learn the \emph{process} for acquiring word embeddings. We believe the language acquisition process is too complex and subtle to handcraft. However, there are large amounts of data available to learn the process.
In this paper, we introduce a framework that \emph{learns how to learn} vision and language representations.

\begin{figure*}[t!]
\includegraphics[width=\textwidth]{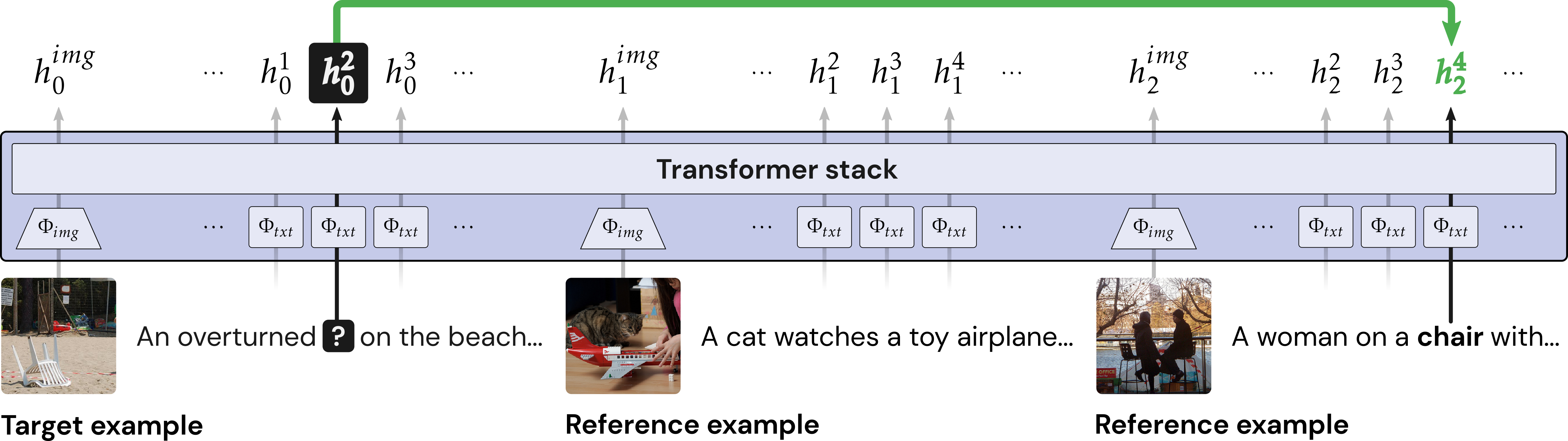}
\caption{\textbf{Learning to Learn Words from Scenes:} Rather than directly learning word embeddings, we instead learn the \emph{process} to acquire word embeddings. The input to our model is an episode of image and language pairs, and our approach meta-learns a policy to acquire word representations from the episode. Experiments show this produces a representation that is able to acquire novel words at inference time as well as more robustly generalize to novel compositions.}
\label{fig:mainidea} 
\end{figure*}

We present a model that receives an episode of examples consisting of vision and language pairs, where the model meta-learns word embeddings from the episode. The model is trained to complete a masked word task, however it must do so by copying and pasting words across examples within the episode. Although this is a roundabout way to fill in masked words, this requires the model to learn a robust process for word acquisition. By controlling the types of episodes from which the model learns, we are able to explicitly learn a process to acquire novel words and generalize to novel compositions.  Figure \ref{fig:mainidea} illustrates our approach. 

Our experiments show that our framework meta-learns a strong policy for word acquisition. We evaluate our approach on two established datasets, Flickr30k \cite{flickr30k} and EPIC-Kitchens  \cite{Damen_2018_ECCV}, both of which have a large diversity of natural scenes and a long-tail word distribution. After learning the policy, the model can receive a stream of images and corresponding short phrases containing unfamiliar words. Our model is able to learn the novel words and point to them to describe other scenes. Visualizations of the model suggest strong cross-modal interaction from language to visual inputs and vice versa.

A key advantage of our approach is that it is able to acquire words with orders of magnitude less examples than previous approaches. Although we train our model from scratch without any language pre-training, it either outperforms or matches methods with massive corpora. In addition, the model is able to effectively generalize to compositions outside of the training set, \emph{e.g.} to unseen compositions of nouns and verbs, outperforming the state-of-the-art in visual language models by over fifteen percent when the compositions are new.

Our primary contribution is a framework that meta-learns a policy for visually grounded language acquisition, which is able to robustly generalize to both new words and compositions. The remainder of the paper is organized around this contribution. In Section~\ref{section:related}, we review related work. In Section~\ref{section:learning}, we present our approach to meta-learn words from visual episodes. In Section~\ref{section:experiments}, we analyze the performance of our approach and ablate components with a set of qualitative and quantitative experiments. We will release all code and trained models.

\section{Related Work}
\label{section:related}

\textbf{Visual language modeling:} Machine learning models have leveraged large text datasets to create strong language models that achieve state-of-the-art results on a variety of tasks \cite{bert,elmo,gpt}. To improve the representation, a series of papers have tightly integrated vision as well \cite{Li2019UnicoderVLAU,Su2019a,Luc,Rahman2019,Li2019b,Zhou2019,Tan2019,Chen2019,Sun2019,Sun2019a,Alberti2019a}. However, since these approaches directly learn the embedding, they often require large amounts of data, poorly generalize to new compositions, and cannot adapt to an open-world vocabulary. In this paper, we introduce a meta-learning framework that instead learns the language acquisition process itself. Our approach outperforms established vision and language models by a significant margin.
Since our goal is word acquisition, we evaluate both our method and baselines on language modeling directly.

\textbf{Compositional models:} Due to the diversity of the visual world, there has been extensive work in computer vision on learning compositional representations for objects and attributes \cite{nagarajan2018attrop,CLEVR,misra2017red,Nikolaus} as well as for objects and actions \cite{Kato,Nikolaus,wray2019fine}. Compositions have also been studied in natural language processing \cite{dasgupta2018evaluating,ettinger2018assessing}. Our paper builds on this foundation. The most related is \cite{LakeCompositional}, which also develops a meta-learning framework for compositional generalization. However, unlike \cite{LakeCompositional}, our approach works for realistic language and natural images. 

\textbf{Out-of-vocabulary words:} This paper is related but different to models of out-of-vocabulary words (OOV) \cite{schick2019rare,lazaridou2017multimodal,herbelot2017high,khodak2018carte,schick2019learning,schick2019attentive,hu2019few,schick2019rare}. Unlike this paper, most of them require extra training, or gradient updates on new words. We compare to the most competitive approach \cite{schick2019rare}, which reduces to regular BERT in our setting, as a baseline. Moreover, we incorporate OOV words not just as an input to the system, but also as output. Previous work on captioning \cite{li2019pointing,lu2018neural,yao2017incorporating,wu2018decoupled} produces words never seen in the ground truth captions. However, they use pretrained object recognition systems to obtain labels and use them to caption the new words. Our paper is different because we instead learn the word acquisition process from vision and text data. Finally, unlike \cite{anne2016deep}, our approach does not require any side information or external information, and instead acquires new words using their surrounding textual and visual context.

\textbf{Few-shot learning:} Our paper builds on foundational work in few-shot learning, which aims to generalize with little or no labeled data. Past work has explored a variety of tasks, including image classification \cite{Xian_2017_CVPR,Sung_2018_CVPR,prototypical}, translating between a language pair never seen explicitly during training \cite{google_translation} or understanding text from a completely new language \cite{Artetxe2019,adams-etal-2017-cross}, among others. In contrast, our approach is designed to acquire \emph{language} from minimal examples. Moreover, our approach is not limited to just few-shot learning. Our method also learns a more robust underlying representation, such as for compositional generalization. 

\textbf{Learning to learn:} Meta-learning is a rapidly growing area of investigation. Different approaches include learning to quickly learn new tasks by finding a good initialization \cite{MAML,li2017meta}, learning efficient optimization policies \cite{schmidhuber:1987:srl,bengio1992optimization,andrychowicz2016learning,ravi2016optimization,li2017meta}, learning to select the correct policy or oracle in what is also known as hierarchical learning \cite{frans2017meta,hu2019few}, and others \cite{duan2016rl,mishra2017simple}. In this paper, we apply meta-learning to acquire new words and compositions from visual scenes. 

\definecolor{dark_color}{RGB}{97,97,97}
\definecolor{light_color}{RGB}{167,167,167}
\begin{figure*}[t]
\centering
\includegraphics[width=\linewidth]{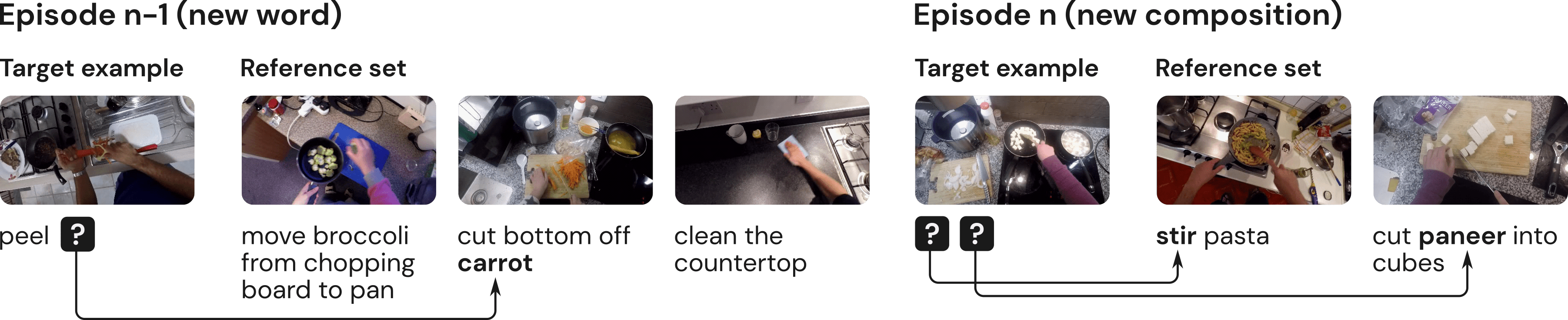}
\caption{\textbf{Episodes for Meta-Learning}: We illustrate two examples of training episodes. Each episode consists of several pairs of image and text. During learning, we mask out one or more words, indicated by a {\setlength{\fboxsep}{2pt}\colorbox{black}{\textcolor{white}{?}}}, and train the model to reconstruct it by pointing to ground truth (in \textbf{bold}) among other examples within the episode. By controlling the generalization gaps within an episode, we can explicitly train the model to generalize and learn new words and new compositions. For example, the left episode requires the model to learn how to acquire a new word (``carrot''), and the right episode requires the model to combine known words to form a novel composition (``stir paneer'').}
\label{fig:eplearn} 
\end{figure*}
\section{Learning to Learn Words}
\label{section:learning}

We present a framework that learns how to acquire words from visual context. In this section, we formulate the problem as a meta-learning task and propose a model that leverages self-attention based transformers to learn from episodes. 

\subsection{Episodes}
\label{subsection:episodes}

We aim to learn the word acquisition process. Our key insight is that we can construct training episodes that demonstrate language acquisition, which provides the data to meta-learn this process. We create training \emph{episodes}, each of which contain multiple \emph{examples} of text-image pairs. During meta-learning, we sample episodes and train the model to acquire words from examples within each episode. Figure~\ref{fig:eplearn} illustrates some episodes and their constituent examples.

To build an episode, we first sample a {\em target example}, which is an image and text pair, and mask some of its word tokens. We then sample {\em reference examples}, some of which contain tokens masked in the target.
We build episodes that require overcoming substantial generalization gaps, allowing us to explicitly meta-learn the model to acquire robust word representations. Some episodes may contain new words, requiring the model to learn a policy for acquiring the word from reference examples and using it to describe the target scene in the episode. Other episodes may contain familiar words but novel compositions in the target. In both cases, the model will need to generalize to target examples by using the reference examples in the episode. Since we train our model on a distribution of episodes instead of a distribution of examples, and each episode contains new scenes, words, and compositions, the learned policy will be robust at generalizing to testing episodes from the same distribution. By propagating the gradient from the target scene back to other examples in the episode, we can directly train the model to learn a word acquisition process. 

\subsection{Model}
\label{subsection:model}

Let an episode be the set $e_k = \{v_1, \ldots, v_i, w_{i+1}, \ldots, w_j\}$ where $v_i$ is an image and $w_i$ is a word token in the episode. We present a model that receives an episode $e_k$, and train the model to reconstruct one or more masked words $w_i$ by pointing to other examples within the same episode. Since the model must predict a masked word by drawing upon other examples within the same episode, it will learn a policy to acquire words from one example and use them for another example. 

\textbf{Transformers on Episodes:} To parameterize our model, we need a representation that is able to capture pairwise relationships between each example in the episode. We propose to use a stack of transformers based on self-attention \cite{Vaswani2017}, which is able to receive multiple image and text pairs, and learn rich contextual outputs for each input \cite{bert}. The input to the model is the episode $\{v_1, \ldots, w_j\}$, and the stack of transformers will produce hidden representations $\{h_1, \ldots, h_j\}$ for each image and word in the episode. 

\textbf{Transformer Architecture:}
We input each image and word into the transformer stack.
One transformer consists of a multi-head attention block followed by a linear projection, which outputs a hidden representation at each location, and is passed in series to the next transformer layer.  Let $H^{z} \in \mathbb{R}^{d \times j}$ be the $d$ dimensional hidden vectors at layer $z$. The transformer first computes vectors for queries $Q = W_q^{z} H^{z}$, keys $K = W_k^{z} H^{z}$, and values $V = W_v^{t} H^{z}$ where each $W_* \in \mathbb{R}^{d\times d}$ is a matrix of learned parameters. Using these queries, keys, and values, the transformer computes the next layer representation by attending to all elements in the previous layer:
\begin{equation}
 H^{z+1} = SV \quad \textrm{where} \quad S = \text{softmax}\left(\frac{QK^T}{\sqrt{d}}\right).
\label{eq:equation_transformer}
\end{equation}
In practice, the transformer uses multi-head attention, which repeats Equation \ref{eq:equation_transformer} once for each head, and concatenates the results.  
The network produces a final representation $\{h_1^Z, \ldots, h_i^Z\}$ for a stack of $Z$ transformers.

\textbf{Input Encoding:}
Before inputting each word and image into the transformer, we encode them with a fixed-length vector representation. To embed input words, we use an $N \times d$ word embedding matrix $\phi_w$, where $N$ is the size of the vocabulary considered by the tokenizer. To embed visual regions, we use a convolutional network $\phi_v(\cdot)$ over images. We use ResNet-18 initialized on ImageNet \cite{russakovsky2015imagenet,he2016deep}. Visual regions can be the entire image in addition to any region proposals. Note that the region proposals only contain spatial information without any category information.

To augment the input encoding with both information about the modality and the positional information (word index for text, relative position of region proposal), we translate the encoding by a learned vector:
\begin{equation}
\begin{aligned}
\phi_{{\textrm{img}}}(v_i) &= \phi_v(v_i) + \phi_{\textrm{loc}}(v_i) + \phi_{\textrm{mod}}(\text{IMG}) + \phi_{\textrm{id}}(v_i) \\
\phi_{\textrm{txt}}(w_j) &= {\phi_w}_j + \phi_{\textrm{pos}}(w_j) + \phi_{\textrm{mod}}(\text{TXT}) + \phi_{\textrm{id}}(w_j)
\end{aligned}
\end{equation}
where $\phi_{\textrm{loc}}$ encodes the spatial position of $v_i$, $\phi_{\textrm{pos}}$ encodes the word position of $w_j$, $\phi_{\textrm{mod}}$ encodes the modality and $\phi_{\textrm{id}}$ encodes the example index.

Please see the Appendix~\ref{impl} for all implementation details of the model architecture.  Code will be released.

\subsection{Learning Objectives}
\label{subsection:objectives}
To train the model, we mask input elements from the episode, and train the model to reconstruct them. We use three different complementary loss terms.

\textbf{Pointing to Words:} 
We train the model to ``point'' \cite{vinyals2015pointer} to other words within the same episode. Let $w_i$ be the target word that we wish to predict, which is masked out. Furthermore, let $w_{i'}$ be the same word which appears in a reference example in the episode ($i' \ne i$). To fill in the masked position $w_i$, we would like the model to point to $w_{i'}$, and not any other word in the reference set.

We estimate similarity between the $i$th element and the $j$th element in the episode.
Pointing to the right word within the episode corresponds to maximizing the similarity between the masked position and the true reference position, which we implement as a cross-entropy loss:
\begin{align}
    \mathcal{L}_{\textrm{point}} = -\log\left(\frac{A_{ii'}}{\sum_k A_{ik}} \right) \quad \textrm{where} \quad  \log A_{ij} = f(h_i)^T f(h_j)
\end{align}
where $A$ is the similarity matrix and $f(h_i) \in \mathbb{R}^d$ is a linear projection of the hidden representation for the $i$th element.
Minimizing the above loss over a large number of episodes will cause the neural network to produce a policy such that a novel reference word $w_{i'}$ is correctly routed to the right position in the target example within the episode. 

Other similarity matrices are possible. The similarity matrix $A$ will cause the model to fill in a masked word by pointing to another contextual representation. However, we can also define a similarity matrix that points to the input word embedding instead. To do this, the matrix is defined as $\log A_{ij} = f(h_i)^T {\phi_w}_j$. This prevents the model from solely relying on the context and forces it to specifically attend to the reference word, which our experiments will show helps generalizing to new words.

\textbf{Word Cloze:} We additionally train the model to reconstruct words by directly predicting them. Given the contextual representation of the masked word $h_i$, the model predicts the missing word by multiplying its contextual representation with the word embedding matrix, $\hat{w_i} = \argmax \phi_w^T h_i$. We then train with cross-entropy loss between the predicted word $\hat{w_i}$ and true word $w_i$, which we write as $\mathcal{L}_{\textrm{cloze}}$. This objective is the same as in the original BERT \cite{bert}.

\textbf{Visual Cloze:} In addition to training the word representations, we train the visual representations on a cloze task. However, whereas the word cloze task requires predicting the missing word, generating missing pixels is challenging. Instead, we impose a metric loss such that a linear projection of $h_{i}$ is closer to $\phi_v(v_i)$ than $\phi_v(v_{k \neq i})$. We use the tripet loss \cite{weinberger2009distance} with cosine similarity and a margin of one. We write this loss as $\mathcal{L}_{\textrm{vision}}$. This loss is similar to the visual loss used in state-of-the-art visual language models \cite{Chen2019}. 

\textbf{Combination:} Since each objective is complementary, we train the model by optimizing the neural network parameters to minimize the sum of losses:
\begin{align}
    \min_\Omega \; \mathbb{E}\left[\mathcal{L}_{\textrm{point}} + \alpha\mathcal{L}_{\textrm{cloze}} + \beta\mathcal{L}_{\textrm{vision}}\right] 
\end{align}
where $\alpha \in \mathbb{R}$ and $\beta \in \mathbb{R}$ are scalar hyper-parameters to balance each loss term, and $\Omega$ are all the learned parameters. We sample an episode, compute the gradients with back-propagation, and update the model parameters by stochastic gradient descent. 

\begin{wrapfigure}{r}{.4\textwidth}
\centering
\includegraphics[width=\linewidth]{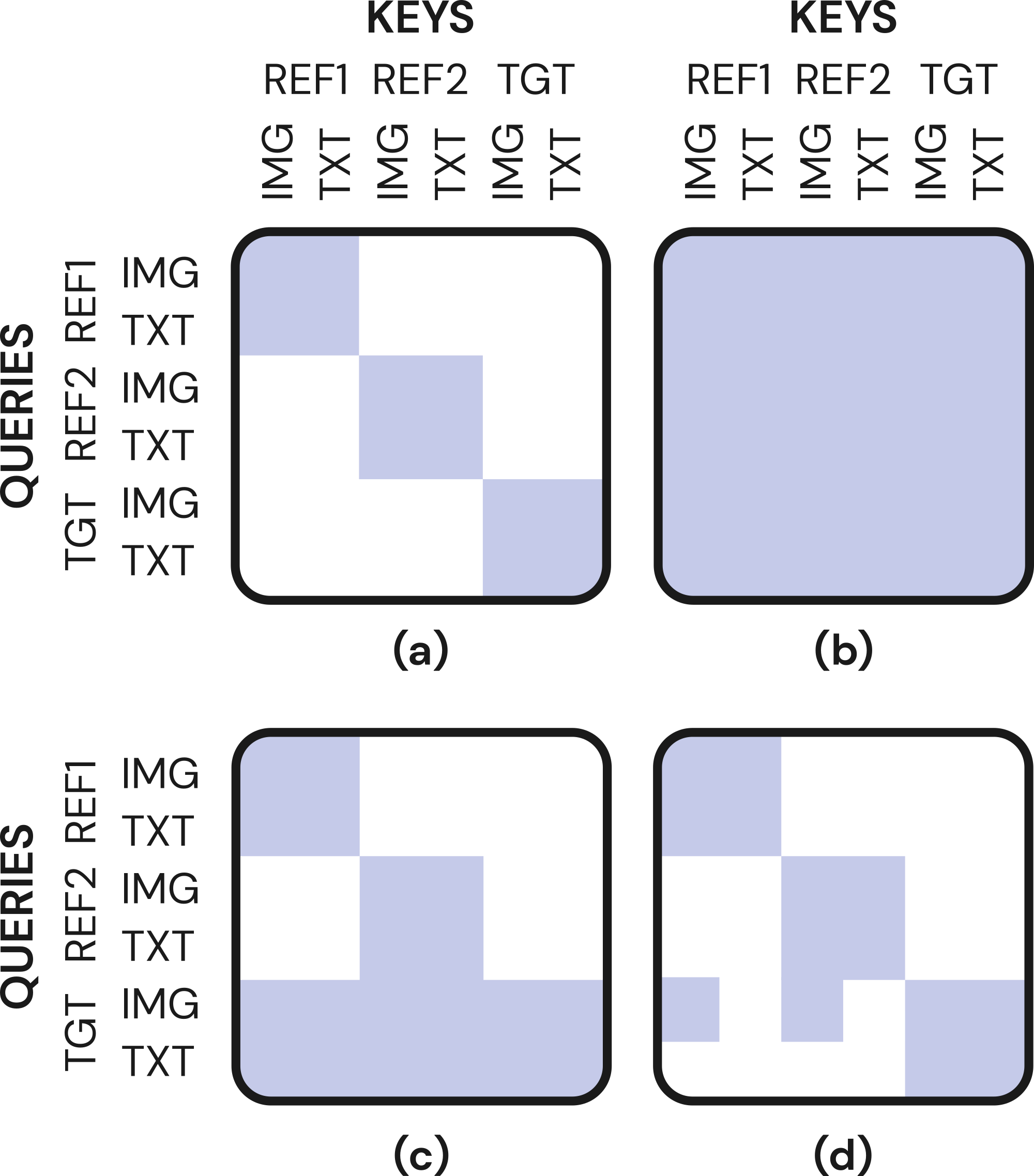}
\caption{We visualize some attention masks for how information can flow through our model, where white locations are masked.}
\label{fig:attention} 
\end{wrapfigure}

\subsection{Information Flow}

To use the episode, information needs to flow from reference examples to the target example. Since the transformer computes attention between elements, we can control how information flows in the model by constraining the attention.  We implement this as a mask on the attention: $H^{z+1} = (S \odot M) V$
where $M_{ij}$ is a binary mask to indicate whether information can flow from element $j$ to $i$. Several masks $M$ are possible. Figure \ref{fig:attention} visualizes them.

\textbf{(a) Isolated attention:} By setting $M_{ij} = 1$ iff $i$ and $j$ belong to the same example in the episode, examples can only attend within themselves. This is equivalent to running each example separately through the model, and optimizing the model with a metric learning loss. 

\textbf{(b) Full attention:} By unconditionally setting $M_{ij} = 1$, attention is fully connected and every element can attend to all other elements. 

\textbf{(c) Target-to-reference attention:} We can constrain the attention to only allow the target elements to attend to the reference elements, and prevent the reference elements from communicating across each other. To do this, $M_{ij} = 1$ iff $i$ and $j$ are from the same example or $i$ is a target element. 

\textbf{(d) Attention via vision:} We can also constrain the attention to only transfer information through vision. Here, $M_{ij} = 1$ if $i$ and $j$ are from the same example, and also $M_{ij} = 1$ if $i$ and $j$ are both images and $i$ is a target element. Otherwise, $M_{ij} = 0$. Information is first propagated from reference text nodes to visual nodes, then propagated from the visual nodes to the target text node. 

Some attention mechanisms are more computationally efficient because they do not require computing representations for all pairwise relationships. For full attention, computation scales quadratically with the number of examples. However, for the other attention mechanisms, computation scales linearly with the number of examples, allowing us to efficiently operate on large episodes. 

\subsection{Inference}
\label{subsection:inference}

After learning, we obtain a policy that can acquire words from an episode consisting of vision and language pairs. Since the model produces words by pointing to them, which is a non-parametric mechanism, the model is consequently able to acquire words that were absent from the training set. As image and text pairs are encountered, they are simply inserted into the reference set. When we ultimately input a target example, the model is able to use new words to describe it by pulling from other examples in the reference set.

Moreover, the model is not restricted to only producing words from the reference set. Since the model is also trained on a cloze task, the underlying model is able to perform any standard language modeling task. In this setting, we only give the model a target example without a reference set. As our experiments will show, the meta-learning objective also improves these language modeling tasks. 

\section{Experiments}
\label{section:experiments}

The goal of our experiments is to analyze the language acquisition process that is learned by our model.  Therefore, we train the model on vision-and-language datasets, without any language pretraining. We call our approach \textbf{EXPERT}.\protect\footnote{Episodic Cross-Modal Pointing for Encoder Representations from Transformers}

\subsection{Datasets}

We use two datasets with natural images and realistic textual descriptions.

\textbf{EPIC-Kitchens} is a large dataset consisting of $39,594$ video clips across $32$ homes. Each clip has a short text narration, which spans $314$ verbs and $678$ nouns, as well as other word types.
EPIC-Kitchens is challenging due to the complexities of unscripted video. We use object region proposals on EPIC-Kitchens, but discard any class labels for image regions. We sample frames from videos and feed them to our models along with the corresponding narration. Since we aim to analyze generalization in language acquisition, we create a train-test split such that some words and compositions will only appear at test time. We list the full train-test split in the Appendix~\ref{epic_split}.

\textbf{Flickr30k} contains $31,600$ images with five descriptions each. The language in Flickr30k is more varied and syntactically complex than in EPIC-Kitchens, but comes from manual descriptive annotations rather than incidental speech.
Images in Flickr30k are not frames from a video, so they do not present the same amount of visual challenges in motion blur, clutter, etc., but they cover a wider range of scene and object categories. We again use region proposals without their labels and create a train-test split that withholds some words and compositions.

Our approach does not require additional image regions as input beyond the full image, and our experiments show that our method outperforms baselines similarly even when trained only with the full image as input, without other cropped regions (see Appendix~\ref{impl}).

\subsection{Baselines}

We compare to established, state-of-the-art models in vision and language, as well as to ablated versions of our approach.

\textbf{BERT} is a language model that recently obtained state-of-the-art performance across several natural language processing tasks \cite{bert}. We consider two variants. Firstly, we download the pre-trained model, which is trained on three billion words, then fine-tune it on our training set. Secondly, we train BERT from scratch on our data. We use BERT as a strong language-only baseline.

\textbf{BERT+Vision} refers to the family of visually grounded language models \cite{Li2019UnicoderVLAU,Alberti2019a,Sun2019a,Zhou2019,Li2019b,Su2019a,nangia2019human,Chen2019}, which adds visual pre-training to BERT. We experimented with several of them on our tasks, and we report the one that performs the best \cite{Chen2019}. Same as our model, this baseline does not use language pretraining.

\begin{SCtable}[1][t]
\centering
 \resizebox{0.5\linewidth}{!}{
\setlength{\tabcolsep}{3pt}
\begin{tabular}{ c l r r r r}
\toprule
 && \multicolumn{2}{c}{\textbf{Ratio}} &     \\
 \cmidrule(lr){3-4}
 &&\textbf{1:1}&\textbf{2:1}& \multicolumn{1}{c}{\makecell{\textbf{Cost}}} \\ \midrule
& \multicolumn{1}{l|}{Chance}             & 13.5 &  8.7 & \multirow{4}{*}{-}  \\ 
& \multicolumn{1}{l|}{BERT (scratch) \cite{bert}}             &  36.5  & 26.3  &  \\ 
& \multicolumn{1}{l|}{BERT+Vision \cite{Chen2019} }   &  63.4  & 57.5  & \\
\midrule
\multirow{5}{*}{\rotatebox[origin=c]{90}{EXPERT}} & 
\multicolumn{1}{l|}{Isolated attention} &  69.0  & 57.8   &$O(n)$ \\
&\multicolumn{1}{l|}{Tgt-to-ref attention}                     &  71.0  & 63.2 &$O(n)$\\ 
&\multicolumn{1}{l|}{Via-vision attention}                      &  72.7  & 64.5  &$O(n)$\\ 
&\multicolumn{1}{l|}{\hspace{0.25cm}+ Input pointing}                     &  \textbf{75.0}  & \textbf{67.4}  &$O(n)$\\ 
&\multicolumn{1}{l|}{Full attention}                      &  \textbf{76.6}  & \textbf{68.4}  & $O(n^2)$  \\ \midrule
& \multicolumn{1}{l|}{BERT (pretrained) \cite{bert}}             &  53.4  & 48.8  &  \\ 
\bottomrule
\end{tabular}

}
\caption{\textbf{Acquiring New Words on EPIC-Kitchens:} We test our model's ability to acquire new words at test time by pointing. The difficulty of this task varies with the number of distractor examples in the reference set. We show \textbf{top-$\mathbf{1}$ accuracy} results on both 1:1 and 2:1 ratios of distractors to positives. The rightmost column shows computational cost of the attention variant used.}
\label{tbl:pting} 
\vspace{-1em}
\end{SCtable}

We also compare several different attention mechanisms. \textbf{Tgt-to-ref attention}, \textbf{Via-vision attention}, and \textbf{Full attention} indicate the choice of attention mask; the base one is the \textbf{Isolated attention}. \textbf{Input pointing} indicates the choice of pointing to the input encodings in addition to contextual encodings. Unless otherwise noted, EXPERT refers to the variant trained with via-vision attention.

\subsection{Acquisition of New Words}

Our model learns the word acquisition \emph{process}. We evaluate this learned process at how well it acquires new words not encountered in the training set.
At test time, we feed the model an episode containing many examples, which contain previously unseen words. Our model has learned a strong word acquisition policy if it can learn a representation for the new words, and correctly use them to fill in the right masked words in the target example. 

Specifically, we pass each example in an episode forward through the model and store hidden representations at each location. We then compute hidden representation similarity between the masked location in the target example and every example in the reference set. We experimented with a few similarity metrics, and found dot-product similarity performs the best, as it is a natural extension of the attention mechanism that transformers are composed of. 

We compare our meta-learned representations to state-of-the-art vision and language representations, i.e.\ BERT and BERT with Vision. When testing, baselines use the same pointing mechanism (similarity score between hidden representations) and reference set as our model. Baselines achieve strong performance since they are trained to learn contextual representations that have meaningful similarities under the same dot-product metric used in our evaluation.  

\begin{figure*}[t!]
\includegraphics[width=\linewidth]{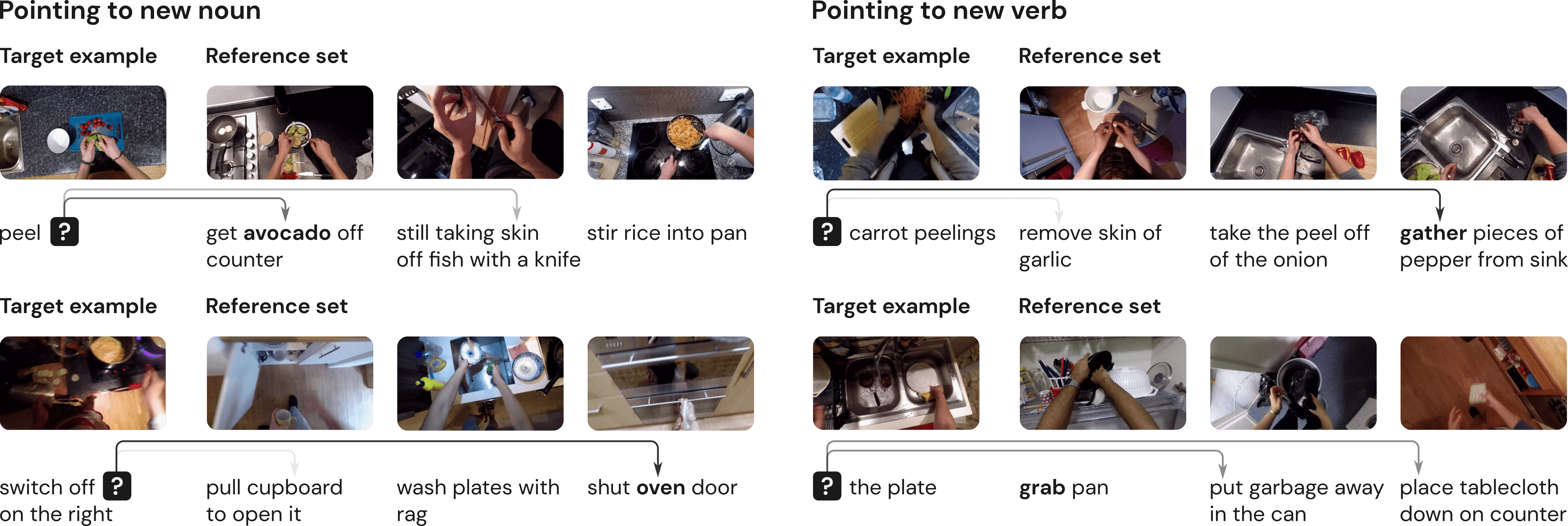}
\caption{\textbf{Word Acquisition:} We show examples where the model acquires new words.
{\setlength{\fboxsep}{2pt}\colorbox{black}{\textcolor{white}{?}}} in the target example indicates the masked out new word. \textbf{Bold} words in the reference set are ground truth. The model makes predictions by pointing into the reference set, and the weight of each pointer is visualized by the shade of the arrows shown (weight $< 3\%$ is omitted). In bottom right, we show an error where the model predicts the plate is being placed, where the ground truth is ``grabbed".}
\label{fig:pting_new} 
\end{figure*}
\tikzset{every picture/.style={/utils/exec={}}}
\usetikzlibrary{arrows}
\pgfplotsset{every tick label/.style={/utils/exec={}}}
\pgfplotsset{compat=1.12}
\begin{SCfigure}[1][t]
\begin{tikzpicture}[scale=0.74]

\begin{axis}[
    xlabel={Distractor Ratio},
    ylabel={Pointing accuracy (\%)},
    xmin=1, xmax=10,
    ymin=20, ymax=80,
    xtick={1,2,3,4,5,6,7,8,9,10},
    ytick={20,30,40,50,60,70,80},
    legend pos=north east,
    ymajorgrids=true,
    grid style=dashed,
    unit vector ratio*=15 1 1,
    legend style={nodes={scale=0.7, transform shape}},
    ticklabel style = {font=\small},
    label style = {font=\small},
    colormap/cool,
]

\definecolor{mycolor1}{RGB}{57,106,177}
\definecolor{mycolor2}{RGB}{218,124,48}
\definecolor{mycolor3}{RGB}{62,150,81}
\definecolor{mycolor4}{RGB}{204,37,41}
\definecolor{mycolor5}{RGB}{83,81,84}
\definecolor{mycolor6}{RGB}{107,76,154}
\definecolor{mycolor7}{RGB}{146,36,40}
\definecolor{mycolor8}{RGB}{148,139,61}

 \addplot+[mycolor1, mark options={fill=mycolor1}]
    coordinates {
    ((1, 75)	(2, 67.4)	(3, 60.4)	(4, 56.3)	(5,52.6)	(6, 48.7)	(7, 46.9)	(8, 44.5)	(9, 43.3)	(10, 41.5)
    }; 
\addplot+[mycolor4, mark options={fill=mycolor4}]
    coordinates {
    (1, 63.4)	(2, 57.5)	(3, 51.1)	(4, 47.8)	(5,42.3)	(6, 40.2)	(7, 37.0)	(8, 33.7)	(9, 32.5)	(10, 30.6)
    };

 \addplot+[mycolor3, mark=triangle*, mark options={fill=mycolor3}]
     coordinates {
(1, 53.4) (2, 48.8)	(3, 45.8)	(4, 44.7)	(5, 39.8)	(6, 36.8)	(7, 34.2)	(8, 32.8)	(9, 30.9)	(10, 29.4)
     };
     
    \legend{EXPERT,BERT+Vision, BERT}

\end{axis}

\end{tikzpicture}
\vspace{-2em}
\caption{\textbf{Word Acquisition versus Distractors:} As more distractors are added (testing on EPIC-Kitchens), the problem becomes more difficult, causing performance for all models to go down. However, EXPERT decreases at a lower rate than baselines.}
\label{fig:negatives_curve}
\end{SCfigure}
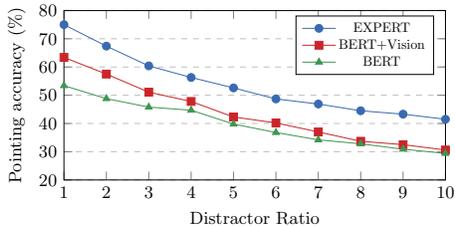
  
We show results on this experiment in Table \ref{tbl:pting}. Our complete model obtains the best performance in word acquisition on both EPIC-Kitchens and Flickr30k. In the case of EPIC-Kitchens, where linguistic information is scarce and sentence structure simpler, meta-learning a strong lexical acquisition policy is particularly important for learning new words. Our model outperforms the strongest baselines (including those pretrained on enormous text corpora) by up to 13\% in this setting. Isolating attention to be only within examples in an episode harms accuracy significantly, suggesting that the interaction between examples is key for performance. Moreover, by constraining this interaction to pass through the visual modality, the computational cost is linear in number of examples with only a minor drop in accuracy. This allows our approach to efficiently scale to episodes with more examples.

Figure~\ref{fig:pting_new} shows qualitative examples where the model must acquire novel language by learning from its reference set, and use it to describe another scene with both nouns and verbs. In the bottom right of the figure, an incorrect example is shown, in which EXPERT points to \textit{place} and \textit{put} instead of \textit{grab}. However, both incorrect options are plausible guesses given only the static image and textual context ``plate''. This example suggests that video information would further improve EXPERT's performance.
  
Figure~\ref{fig:negatives_curve} shows that, even as the size of the reference set (and thus the difficulty of language acquisition) increases, the performance of our model remains relatively strong. EXPERT outperforms baselines by $18\%$ with one distractor example, and by $36\%$ with ten. This shows that our model remains relatively robust compared to baselines.

\begin{SCtable}[2][t]
\centering
\resizebox{0.35\linewidth}{!}{
\begin{tabular}{ l | r r  }
\toprule
 & \multicolumn{2}{c}{\textbf{Ratio}}     \\
 \cmidrule(lr){2-3}

\textbf{Method} & \textbf{1:1} &  \textbf{2:1} \\ \midrule
Chance & 3.4 & 2.3 \\
BERT (scratch) & 31.6 & 25.7 \\
BERT with Vision \cite{Chen2019} & 32.1 & 26.8 \\
EXPERT & \textbf{69.3} & \textbf{60.9} \\ \midrule
BERT (pretrained) & \textbf{69.4} & \textbf{60.8} \\
\bottomrule
\end{tabular}
}
\caption{\textbf{Acquiring New Words on Flickr30k:} We run the same experiment as Table \ref{tbl:pting} (\textbf{top-$\mathbf{1}$ accuracy} pointing to new words), except on the Flickr30k dataset, which has more complex textual data. As before, we show results on 1:1 and 2:1 ratios of distractors to positives. By learning the acquisition policy, our model obtains competitive performance with orders of magnitude less training data.}
\label{tbl:pting_flickr} 
\end{SCtable}

In Flickr30k, visual scenes are manually described in text by annotators rather than transcribed from incidental speech, so they present a significant challenge in their complexity of syntactic structure and diversity of subject matter. In this setting, our model significantly outperforms all baselines that train from scratch on Flickr30k, with an increase in accuracy of up to 37\% (Table \ref{tbl:pting_flickr}). Since text is more prominent, a state-of-the-art language model pretrained on huge ($>3$ billion token) text datasets performs well, but EXPERT achieves the same accuracy while requiring several orders of magnitude less training data.

\subsection{Acquisition of Familiar Words}
\label{subsection:manyexamples}
  
\begin{SCtable}[1][t]
\resizebox{0.6\linewidth}{!}{
\begin{tabular}{ l | r r r | r r r  }
\toprule
& \multicolumn{3}{c}{\textbf{EPIC-Kitchens}} &  \multicolumn{3}{|c}{\textbf{Flickr30k}} \\ 
\textbf{Method} & \multicolumn{1}{c}{\textbf{Verbs}} &  \multicolumn{1}{c}{\textbf{Nouns}}  &  \multicolumn{1}{c|}{\textbf{All}} & \multicolumn{1}{c}{\textbf{Verbs}} &  \multicolumn{1}{c}{\textbf{Nouns}}  &  \multicolumn{1}{c}{\textbf{All}}  \\ \midrule
Chance & 0.1 & 0.1 & 0.1 & $<0.1$ & $<0.1$ & $<0.1$ \\ 
BERT (scratch) \cite{bert}                    &  68.2  & 48.9  & 57.9 & 64.8 & 69.4 & 66.2 \\ 
BERT with Vision \cite{Chen2019}    &  77.3  & 63.2  & 65.6 & 65.1 & 70.2 & 66.5\\ 
EXPERT           &  \textbf{81.9}  & \textbf{73.0} &  \textbf{74.9} & \textbf{69.1} & \textbf{79.8} & \textbf{72.0} \\ \midrule
BERT (pretrained) \cite{bert}                    &  71.4  & 51.5  & 59.8 & \textbf{69.5} & \textbf{79.4} & \textbf{72.2} \\ 
\bottomrule
\end{tabular}
}
\caption{\textbf{Acquiring Familiar Words:} We report \textbf{top-$\mathbf{5}$ accuracy} on masked language modeling of words which appear in training. Our model outperforms all other baselines.
}
\label{tbl:inst} 
\end{SCtable}

By learning a policy for word acquisition, the model also jointly learns a representation for the familiar words in the training set. Since the representation is trained to facilitate the acquisition process, we expect these embeddings to also be robust at standard language modeling tasks. We directly evaluate them on the standard cloze test \cite{cloze_test}, which all models (including baselines) are trained to complete.

Table \ref{tbl:inst} shows performance on language modeling. The results suggest that visual information helps learn a more robust language model. Moreover, our approach, which learns the process in addition to the embeddings, outperforms all baselines by between 4 and 9 percent across both datasets. While a fully pretrained BERT model also obtains strong performance on Flickr30k, our model  is able to match its accuracy with orders of magnitude less training data.w

Our results suggest that learning a process for word acquisition also collaterally improves standard vision and language modeling. We hypothesize this happens because learning acquisition provides an incentive for the model to generalize, which acts as a regularization for the underlying word embeddings.

\subsection{Compositionality}
\label{subsection:comp}

\begin{SCtable}[2][t]
\resizebox{0.6\linewidth}{!}{
\begin{tabular}{ l c c | c | c c | c  }
\toprule
& \multicolumn{3}{|c}{\textbf{EPIC-Kitchens}} & \multicolumn{3}{|c}{\textbf{Flickr30k}} \\
\cmidrule(lr){2-7}
\textbf{Method} & \multicolumn{1}{|c}{\textbf{Seen}} &  \multicolumn{1}{c}{\textbf{New}} & \multicolumn{1}{|c}{\textbf{Diff}}  & \multicolumn{1}{|c}{\textbf{Seen}} &  \multicolumn{1}{c}{\textbf{New}} & \multicolumn{1}{|c}{\textbf{Diff}} \\ \midrule
\multicolumn{1}{l|}{Chance}                    & $<0.1$ & $<0.1$ & - &$<0.1$ & $<0.1$ & -\\ 
\multicolumn{1}{l|}{BERT (scratch) \cite{bert}}                    &  34.3  & 17.7 & 16.6 & 43.4 & 39.4 & 4.0\\ 
\multicolumn{1}{l|}{BERT with Vision \cite{Chen2019}}    &  56.1  & 37.6 & 18.5 & 45.0 & 42.0 & 3.0 \\ 
\multicolumn{1}{l|}{EXPERT}           &  \textbf{63.5}  & \textbf{53.0} & \textbf{10.5} & \textbf{48.7} & \textbf{47.1} & \textbf{1.6} \\ \midrule
\multicolumn{1}{l|}{BERT (pretrained) \cite{bert}}                    &  39.8  & 20.7 & 19.1 & \textbf{48.8} & \textbf{47.2} & \textbf{1.6}\\ 
\bottomrule
\end{tabular}
}
\caption{\textbf{Compositionality:} We show top-$5$ accuracy at predicting masked compositions of seen nouns and verbs. Both the verb and the noun must be correctly predicted. EXPERT achieves the best performance on both datasets.}
\label{tbl:composition}  
\end{SCtable}

Since natural language is compositional, we quantify how well the representations generalize to novel combinations of verbs and nouns that were absent from the training set. We again use the cloze task to evaluate models, but require the model to predict both a verb and a noun instead of only one word.

We report results on compositions in Table \ref{tbl:composition} for both datasets. We breakdown results by whether the compositions were seen or not during training. Note that, for all approaches, there is a substantial performance gap between seen and novel compositions. However, since our model is explicitly trained for generalization, the gap is significantly smaller (nearly twice as small). Moreover, our approach also shows substantial gains over baselines for both seen and novel compositions, improving by seven and sixteen points respectively. Additionally, our approach is able to exceed or match the performance of pretrained BERT, even though our model is trained on three orders of magnitude less training data. 

\subsection{Retrieval}

Following prior work \cite{Luc,Su2019a,Li2019UnicoderVLAU,Chen2019}, we evaluate our representation on two cross-modal retrieval tasks for the two datasets. In all cases, we observe significant gains from our approach, outperforming baselines by up to 19\%. Specifically, we run an image/text cross-modal retrieval test on both the baseline BERT+Vision model and ours. We freeze model weights and train a classifier on top to decide whether input image and text match, randomly replacing data from one modality to create negative pairs. We then test on samples containing new compositions. Please see Table~\ref{table:retrieval} for results.

\begin{SCtable}[2][t]
\resizebox{0.6\linewidth}{!}{
\begin{tabular}{ l  c c  c c }
\toprule
& \multicolumn{2}{|c}{\textbf{EPIC-Kitchens}} & \multicolumn{2}{|c}{\textbf{Flickr30k}} \\

\textbf{Method} & \multicolumn{1}{|c}{T$\rightarrow$I} &  I$\rightarrow$T & \multicolumn{1}{|c}{T$\rightarrow$I} &  I$\rightarrow$T \\ 
\midrule
\multicolumn{1}{l|}{Chance}                    & 10.0 & 10.0 &  \multicolumn{1}{|c}{10.0} & 10.0 \\ 
\multicolumn{1}{l|}{BERT with Vision \cite{Chen2019}}    & 13.8  & 13.9 &  \multicolumn{1}{|c}{54.9} & 57.4  \\ 
\multicolumn{1}{l|}{EXPERT}           &  \textbf{32.6}  & \textbf{25.3} &  \multicolumn{1}{|c}{\textbf{57.5}} & \textbf{60.6} \\
\bottomrule
\end{tabular}
}
\caption{\textbf{Retrieval:} We test the model's top-1 retrieval accuracy (in \%) from a 10 sample retrieval set. T$\rightarrow$I and I$\rightarrow$T represent retrieval from image to text and text to image.}
\label{table:retrieval}  
\end{SCtable}

\subsection{Analysis}
\label{subsection:analysis}

\definecolor{myblue_light}{RGB}{223,230,243}
\definecolor{myblue_dark}{RGB}{75,83,97}
\definecolor{myred_light}{RGB}{244,226,225}
\definecolor{myred_dark}{RGB}{117,96,95}

\begin{figure*}[t!]
\centering
\includegraphics[width=\textwidth]{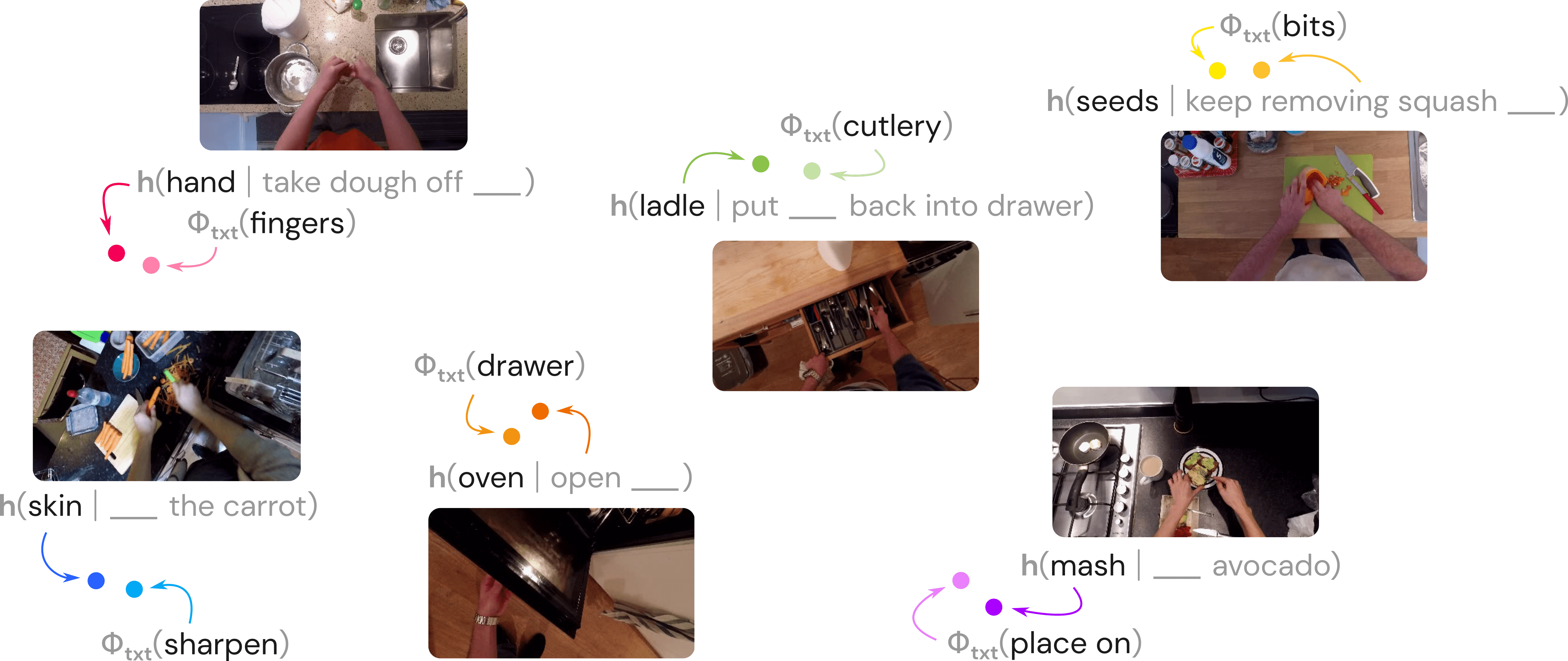}
\caption{\textbf{Embedding New Words with EXPERT:} We give EXPERT sentences with unfamiliar language at test time. We show the hidden vectors $h(\textit{new word} \mid \textit{context}, \textit{image})$ it produces, conditioned on visual and linguistic context, and their nearest neighbors in word embedding space $\phi_{txt}(\textit{known word})$. EXPERT can use its learned vision-and-language policy to embed new words near other words that are similar in object category, affordances, and semantic properties.}
\label{fig:new_concepts} 
\end{figure*}

In this section, we analyze \emph{why} EXPERT obtains better performance.

\textbf{How are new words embedded in EXPERT?} Figure \ref{fig:new_concepts} shows how EXPERT represents new words in its embedding space at test time. We run sentences which contain previously unseen words through our model. Then, we calculate the nearest neighbor of generated hidden representations of these unseen words in the learned word embedding matrix. Our model learns a representation space such that new words are embedded near semantically similar words (dependent on context), even though we use no such supervisory signal in training.

\textbf{Does EXPERT use vision?} We take our complete model, trained with both text and images, and withhold images at test time. Performance drops to nearly chance, showing that EXPERT uses visual information to predict words and disambiguate between similar language contexts. 

\begin{SCfigure}[1][t]
\includegraphics[width=0.45\linewidth]{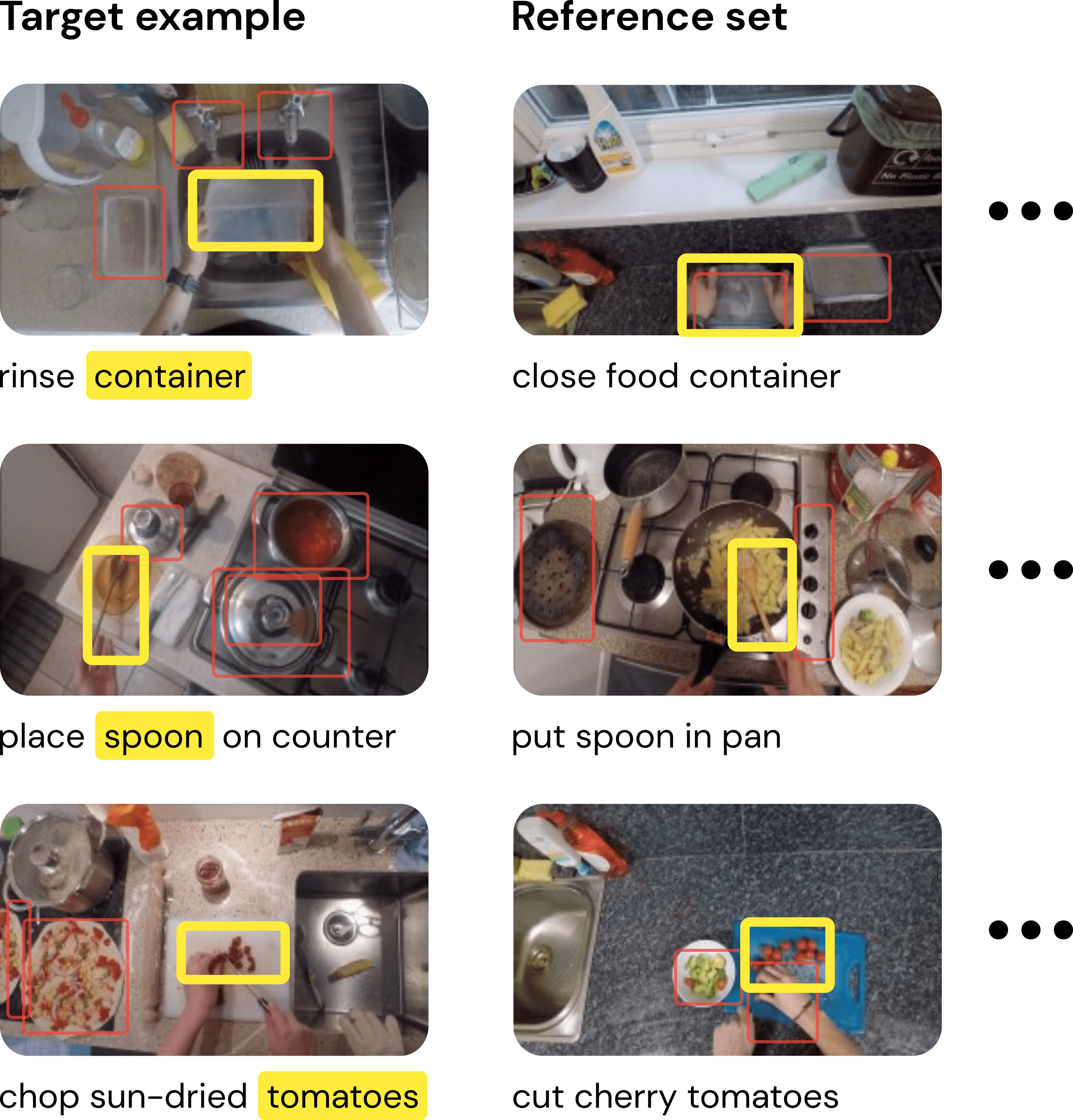}
\caption{\textbf{Visualizing the Attention:} We probe how the model uses visual information. We remove various objects from input images in an episode, and evaluate the model's confidence in predicting
\protect\colorbox[rgb]{1,.9216,.2313}{\textcolor[rgb]{.1,.1,.1}{masked words}}. Removing image regions with a \protect\cfbox{new_green}{\textbf{\textcolor[rgb]{1,.9216,.2313}{yellow box}}} causes the greatest drop in confidence (other regions are shown in \protect\cfbox{new_pink}{\textbf{\textcolor[rgb]{.9569, .2627, .2118}{red}}}). The most important visual regions for the prediction task contain an instance of the target word. These results suggest that our model learns some spatial localization of words automatically.}
\label{fig:attn_inspection}
\end{SCfigure}

\textbf{What visual information does EXPERT use?} To study this, we withhold one visual region at a time from the episode and find the regions that cause the largest decrease in prediction confidence. Figure \ref{fig:attn_inspection} visualizes these regions, showing that removing the object that corresponds to the target word causes the largest drop in performance. This suggests that the model is correlating these words with the right visual region, without direct supervision.

\begin{figure}[t]
    \centering
    \includegraphics[width=0.8\linewidth]{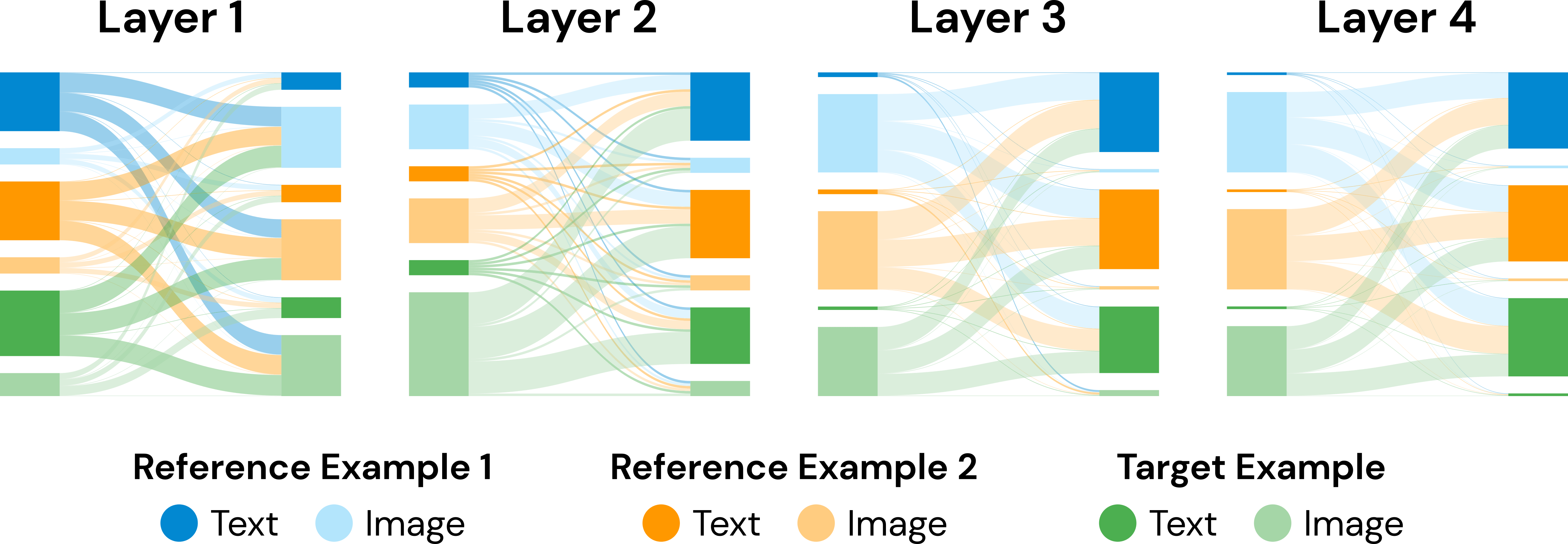}
    \caption{\textbf{Visualizing the Learned Process:} We visualize how information flows through the learned word acquisition process. The width of the pipe indicates the importance of the connection, as estimated by how much performance drops if removed. In the first layer, information tends to flow from the textual nodes to the image nodes. In subsequent layers, information tends to flow from image nodes back to text nodes.}
    \label{fig:attn_abl}
\end{figure}

\textbf{How does information flow through EXPERT?} Our model makes predictions by attending to other elements within its episode. To analyze the learned attention, we take the variant of EXPERT trained with full pairwise attention and measure changes in accuracy as we disable query-key interactions one by one. Figure \ref{fig:attn_abl} shows which connections are most important for performance. This reveals a strong dependence on cross-modal attention, where information flows from text to image in the first layer, and back to text in the last layer.

\textbf{How does EXPERT disambiguate multiple new words?} We evaluate our model on episodes that contain five new words in the reference set, only one of which matches the target token. Our model obtains an accuracy of $56\%$ in this scenario, while randomly picking one of the novel words would give $20\%$. This shows that our model is able to discriminate between many new words in an episode. We also evaluate the fine-tuned BERT model in this same setting, where it obtains a $37\%$ accuracy, significantly worse than our model. This suggests that vision is important to disambiguate new words.

\section{Discussion}

We believe the language acquisition process is too complex to hand-craft. In this paper, we instead propose to meta-learn a policy for word acquisition from visual scenes. Compared to established baselines across two datasets, our experiments show significant gains at acquiring novel words, generalizing to novel compositions, and learning more robust word representations. Visualizations and analysis reveal that the learned policy leverages both the visual scene and linguistic context.

\textbf{Acknowledgements:} We thank Alireza Zareian, Bobby Wu, Spencer Whitehead, Parita Pooj and Boyuan Chen for helpful discussion. Funding for this research was provided by DARPA GAILA HR00111990058. We thank NVidia for GPU donations.


\bibliographystyle{splncs04}
\bibliography{egbib}

\section*{Appendix}
\addcontentsline{toc}{section}{Appendix}
\renewcommand{\thesubsection}{\Alph{subsection}}

\subsection{EPIC-Kitchens Train-Test Split}
\label{epic_split}

The EPIC-Kitchens dataset does not provide action narrations for its test set, which we need for evaluation. We therefore create our own train-test split from the dataset following their conventions. We aim to generate an 80-20 train-test split. We select a small subset of verbs and nouns to remove entirely from the training set, as well as verb-noun compositions. We remove around 4\% of nouns and verbs and withhold them for testing, and around 10\% of all compositions. These are not uniformly distributed, and removing a verb or noun entirely from the training set often results in withholding a disproportionate amount of data. Therefore, with these parameters, we end up with around a 81-19 split. We will release this dataset splits for others to build on our this work.

Below, we list the words and compositions that are withheld from the training set, but present during test.

\hspace{-0.4cm}\begin{minipage}[t][][t]{0.45\linewidth}%
\raggedright%
\textbf{List of new nouns:}
\begin{minipage}[t][][t]{0.5\linewidth}%
\noindent%
\begin{enumerate}[topsep=0pt,itemsep=-2pt,leftmargin=13pt]
    \item avocado
    \item counter
    \item hand
    \item ladle
    \item nesquik
    \item nut
    \item oven
\end{enumerate}
\end{minipage}
\begin{minipage}[t][][t]{0.5\linewidth}
\raggedright
\begin{enumerate}[topsep=0pt,itemsep=-2pt,leftmargin=13pt]
\setcounter{enumi}{7}
    \item peeler
    \item salmon
    \item sandwich
    \item seed
    \item onion
    \item corn
\end{enumerate}
\end{minipage}
\end{minipage}
\hspace{0.3cm}\begin{minipage}[t][][t]{0.5\linewidth}
\raggedright
\textbf{List of new verbs:}
\begin{minipage}[t][][t]{\linewidth}
\noindent
\begin{enumerate}[topsep=0pt,itemsep=-2pt,leftmargin=13pt]
\item fry
\item gather
\item grab
\item mash
\item skin
\item watch
\end{enumerate}
\end{minipage}
\end{minipage}
\\
\vspace{1em}

\hspace{-0.4cm}\begin{minipage}[h]{\linewidth}%
\raggedright
\textbf{List of new verb/noun compositions:}\\
\noindent Since there are around 400 new compositions, we include only the 20 most common here. \\

\begin{minipage}[t][][t]{0.5\linewidth}
\noindent
\begin{enumerate}[topsep=0pt,itemsep=-2pt,leftmargin=13pt]
\item close oven
\item cut peach
\item dry hand
\item fry in pan
\item grab plate
\item open cupboard
\item open oven
\item pick up sponge
\item put onion
\item put in oven
\end{enumerate}
\end{minipage}
\begin{minipage}[t][][t]{0.5\linewidth}
\noindent
\begin{enumerate}[topsep=0pt,itemsep=-2pt,leftmargin=13pt]
\setcounter{enumi}{10}
\item put spoon
\item remove garlic
\item rinse hand
\item skin carrot
\item stir pasta
\item take plate
\item wash hand
\item wash knife
\item wipe counter
\item wipe hand
\end{enumerate}
\end{minipage}
\end{minipage}

 \vspace{0.3cm}
 
 Since the Flickr30k dataset has a larger vocabulary than EPIC-Kitchens and thus a larger train-test split, we only list here the new verbs and nouns, and not the new compositions.
 
\textbf{Flickr30k new nouns:} 3d, a, A\&M, ad, Adidas, Africa, AIDS, airborne, Airways, Alaska, America, american, Americans, Amsterdam, Angeles, Asia, ax, B, badminton, bale, barrier, Batman, batman, Bay, be, Beijing, Berlin, big, Birmingham, Blue, Boston, bottle, Brazil, Bridge, Britain, British, british, Bush, c, California, Calvin, Canada, canadian, Canyon, card, Carolina, case, catholic, cause, cd, cello, Celtics, Chicago, China, chinatown, Chinese, chinese, christmas, Circus, Claus, Clause, clipper, co, cocktail, colleague, Coney, content, convertible, courtyard, Cruz, cup, cycle, dance, David, Deere, desk, Dior, Disney, dj, Domino, dragster, drum, dryer, DS, dump, east, Easter, eastern, Eiffel, Elmo, elmo, England, Europe, fellow, Florida, Ford, France, Francisco, Gate, gi, Giants, glove, go, God, Golden, golden, graduate, Grand, Great, Haiti, halloween, hedge, Heineken, helicopter, hi, hide, highchair, hispanic, Hollister, Hollywood, hoop, Houston, iMac, India, Indian, Indians, information, ingredient, Iowa, Island, Israel, Italy, Jackson, jam, Japan, Jesus, Jim, Joe, John, Klein, knoll, La, Lakers, Las, latex, layup, legged, liberty, library, Lincoln, locomotive, London, loom, Los, Lynyrd, ma, mariachi, Mets, Mexico, Miami, Michael, Michigan, Mickey, mickey, Mike, Miller, mind, Morgan, Mr, Mrs, Music, muslim, Navy, New, new, NFL, Nintendo, no, nun, NY, NYC, o, Obama, officer, Oklahoma, old, op, oriental, ox, Oxford, p, Pabst, Pacific, Paris, Patrick, Paul, pavement, pc, Penn, pew, piercing, pig, plane, pot, punk, rafting, rain, razor, Red, Renaissance, repairman, research, robe, Rodgers, runway, RV, S, Salvation, San, saris, scrimmage, scrubs, Seattle, second, shooting, shrine, Skynyrd, something, South, Sox, Spain, Spanish, Square, SquarePants, St, St, start, States, Statue, Story, style, superman, surfs, T, t, tech, Texas, texas, the, Thomas, Times, today, tooth, Toronto, Toy, trinket, tv, type, U, UFC, UK, ultimate, Unicef, United, up, US, USA, v, Vegas, Verizon, Volvo, vw, W, Wall, Wars, Washington, Wells, West, White, Wii, wind, windsurfer, Winnie, winter, Wonder, wonder, x, Yankees, yard, yo, yong, York, york.

\textbf{Flickr30k new verbs:} address, am, amused, applaud, armed, baked, bar, be, bearded, blend, boat, bound, broken, build, button, cling, complect, confused, cooked, costumed, covered, crashing, crowded, crumble, darkened, decorated, deflated, do, dyed, fallen, fenced, file, flying, go, goggle, graze, haircut, handicapped, inflated, injured, juggle, listening, lit, living, looking, magnify, measure, mixed, motorize, mounted, muzzle, muzzled, numbered, oncoming, opposing, organized, patterned, pierced, populated, proclaim, puzzle, restrain, rim, saddle, scratch, seated, secure, sell, shaved, skinned, slump, solder, spotted, sprawl, streaked, strike, stuffed, suited, sweeping, tanned, tattooed, tile, tiled, train, uniformed, up, wheeled, woode, wooded.
 
\subsection{Language Model Results}
\label{lm}
\begin{figure*}[t]
\includegraphics[width=\linewidth]{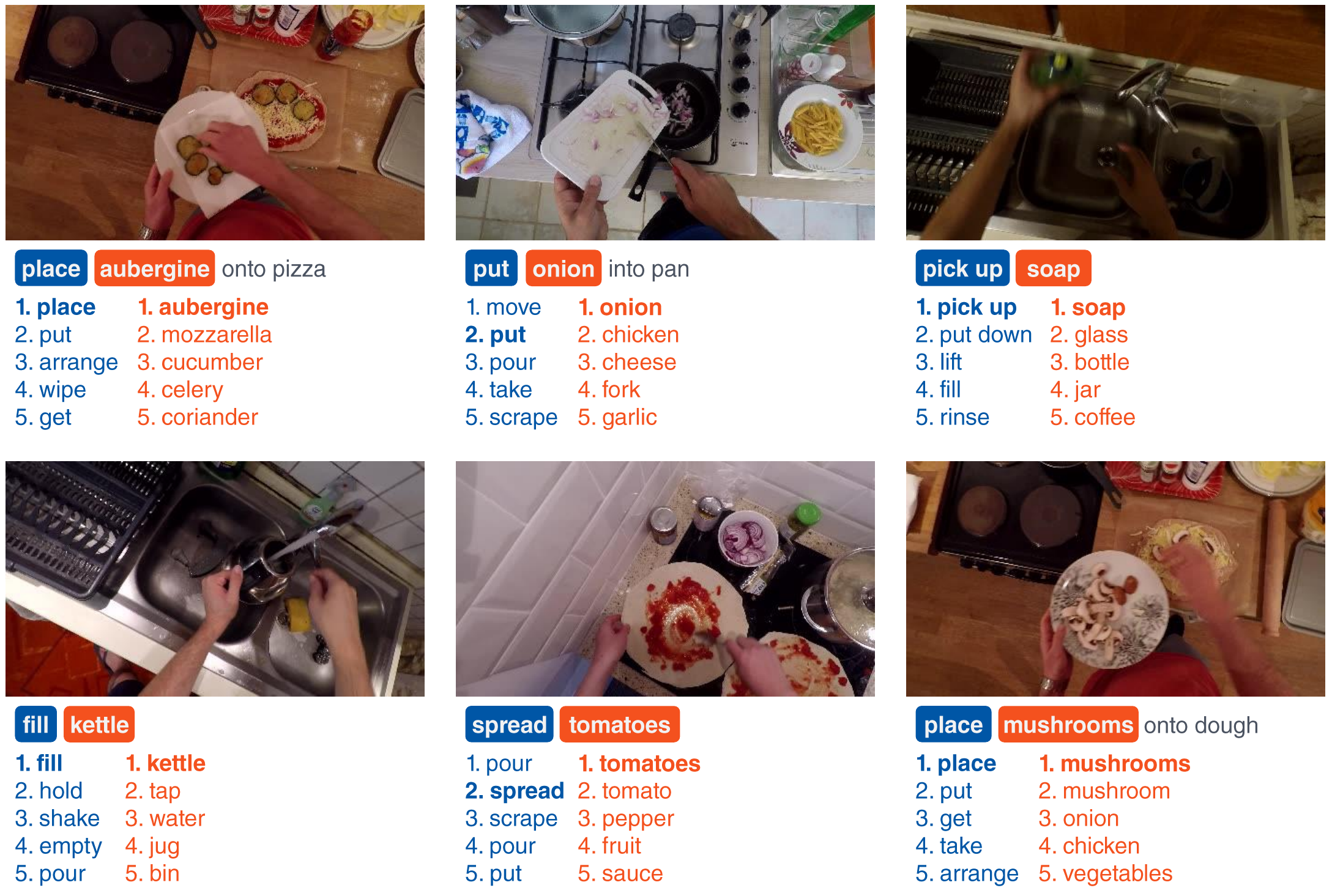}
\caption{\textbf{Predictions of new compositions:} We show some examples of our model's ability to generalize to new compositions, given only the images shown (and their bounding boxes), as well as the unmasked words in the sentence.  We show the top five predictions for each word. As in the paper, we indicate masked words with the {\setlength{\fboxsep}{2pt}\colorbox{black}{\textcolor{white}{dark box}}}, and their predictions below it.}
\label{fig:morelm} 
\end{figure*}

We show EXPERT's outputs when given a sentence containing a new composition of verb and noun. The verb and noun are masked, and we ask EXPERT to make language model predictions at these locations (as in the standard BERT cloze task setting). Results are shown in Figure~\ref{fig:morelm}.\par
In addition, we provide detailed language modeling performance figures in Tables \ref{tbl:inst_app} and \ref{tbl:composition_app}, breaking accuracy down by model variant.

\subsection{Implementation Details}
\label{impl}

For all our experiments we use four transformers ($Z=4$) and four heads. All the models are trained with the Adam optimizer, with a learning rate of $3 \times 10^{-5}$ and $\epsilon=0.0001$. In our experiments, optimization typically takes one week on a single GPU.


In training, we mask out text tokens $\frac{1}{3}$ of the time and image tokens $\frac{1}{6}$ of the time. Following \cite{bert}, a masked text token gets assigned a special \texttt{[MASK]} token 80\% of the time, a random word token 10\%, and remains unchanged 10\%. Similarly, we zero out image tokens 90\% of the time and leave them unaltered 10\%.\par

We construct episodes by first randomly sampling a target example from the training set. We then randomly select between $0$ and $k_+=2$ text tokens as targets, which will be masked with probability $1$. For each one of these tokens, we randomly add to the episode another example in the training set whose text contains the token. We then randomly add between $0$ and $k_{\_}=2$ negative examples (distractors) to the episode, which do not contain any of the target tokens in their text.

We randomly shuffle examples in an episode before feeding them to the model. Then, we combine them with indicator tokens to demarcate examples: \texttt{[IMG]} $v^1_1, \ldots, v^1_{I_1}$ \texttt{[SEP]} \ldots \texttt{[SEP]} $v^k_1, \ldots, v^k_{I_k}$  \texttt{[TXT]} $w^1_1, \ldots, w^1_{J_1}$ \texttt{[SEP]} \ldots \texttt{[SEP]} $w^k_1, \ldots, w^k_{J_k}$ \texttt{[SEP]}. We denote example index with the superscript and $k = \textrm{rand}(0, k_{\_}) + \textrm{rand}(0, k_+) + 1$ is the number of examples in the episode. $I_i$ and $J_i$ are the number of image and text tokens in the i\textsuperscript{th} example of the episode. 

We use the PyTorch framework \cite{paszke2017automatic} to implement our model. We base much of our code on a modified version of the Hugging Face PyTorch transformer repository.~\footnote{https://huggingface.co/transformers/} In particular, we implement our own model class that extends \texttt{BertPreTrainedModel}. We write our own input encoding class that extends \texttt{BertEmbeddings} and adds visual embedding functionality as well as all the $\phi$ functions that are added to the word and image region embeddings. We use \texttt{hidden\_size=384, intermediate\_size=1536, num\_attention\_heads=4, \\num\_hidden\_layers=4, max\_position\_embeddings=512,  type\_vocab\_size=64, vocab\_size=32000} (and all other parameters default) to configure all models except the pretrained BERT one, which uses the original weights from \texttt{BERT-small}.

When we give an example to the model (whether individually or as part of an episode), we include all bounding boxes provided in the EPIC-Kitchens and Flickr30k datasets as well as the whole scene image, though our method can generalize to all formats of image input (\emph{e.g.} multiple whole image frames, to provide temporal information and help action disambiguation). In practice, we filter out all bounding boxes of width or height $<10\text{px}$ since they do not provide useful information to the model, and resize both the entire image and bounding boxes to $112 \times 112$.



\begin{table}[tb]
\centering
\begin{tabular}{c  l | r r r  }
\toprule
&\textbf{Method} & \multicolumn{1}{c}{\textbf{Verbs}} &  \multicolumn{1}{c}{\textbf{Nouns}}  &  \multicolumn{1}{c}{\textbf{All PoS}}  \\ \midrule
&Chance & 0.1 & 0.1 & 0.1 \\ 
&BERT (scratch) \cite{bert}                    &  68.2  & 48.9  & 57.9 \\ 
&BERT (pretrained) \cite{bert}                    &  71.4  & 51.5  & 59.8 \\ 
&BERT with Vision \cite{Chen2019}    &  77.3  & 63.2  & 65.6 \\ 
\midrule
\multirow{5}{*}{\rotatebox[origin=c]{90}{EXPERT}} 
&\multicolumn{1}{l|}{Isolated attn}             &   81.2  & 74.2 &  66.8\\ 
&\multicolumn{1}{l|}{Target-to-ref attn}        &  80.9  & 69.6  & 69.8 \\ 
&\multicolumn{1}{l|}{Via-vision attn}           & 81.9  & 73.0 &  74.9 \\ 
&\multicolumn{1}{l|}{   + Input pointing}       &  80.1 & 73.2   & 67.0 \\ 
&\multicolumn{1}{l|}{Full attn}                 &  79.4 &  68.7 & 67.3  \\ 
\bottomrule
\end{tabular}
\vspace{-0.5em}
\caption{\textbf{Acquiring Familiar Words on EPIC-Kitchens:} We report top-$5$ accuracy at predicting words seen in training for new visual instances.\vspace{-1em}}
\label{tbl:inst_app} 
\end{table}

\begin{table}[t]
\centering
\begin{tabular}{ c l c c | c  }
\toprule
&\textbf{Method} & \multicolumn{1}{|c}{\textbf{Seen}} &  \multicolumn{1}{c}{\textbf{New}} & \multicolumn{1}{|c}{\textbf{Difference}} \\ \midrule
&\multicolumn{1}{l|}{Chance}                    &  \texttildelow  0 & \texttildelow  0 & -\\ 
&\multicolumn{1}{l|}{BERT (scratch) \cite{bert}}                    &  34.3  & 17.7 & 16.6\\ 
&\multicolumn{1}{l|}{BERT (pretrained) \cite{bert}}                    &  39.8  & 20.7 & 19.1\\ 
&\multicolumn{1}{l|}{BERT with Vision \cite{Chen2019}}    &  56.1  & 37.6 & 18.5\\ 
\midrule
\multirow{5}{*}{\rotatebox[origin=c]{90}{EXPERT}} 
&\multicolumn{1}{l|}{Isolated attn}             & 65.0  & 51.6   & 13.4 \\ 
&\multicolumn{1}{l|}{Target-to-ref attn}        & 61.7  & 45.7  & 16.0 \\ 
&\multicolumn{1}{l|}{Via-vision attn}           &  63.5  & 53.0 & 10.5 \\ 
&\multicolumn{1}{l|}{   + Input pointing}       &  62.7 & 48.3  & 15.4 \\ 
&\multicolumn{1}{l|}{Full attn}                 & 59.2  & 44.4  & 14.8 \\ 
\bottomrule
\end{tabular}
\vspace{-0.5em}
\caption{\textbf{Compositionality on EPIC-Kitchens:} We show top-$5$ accuracy at predicting masked compositions of seen nouns and verbs -- both the verb and the noun must be correctly predicted.\vspace{-1em}}
\label{tbl:composition_app}     
\end{table}

\begin{table}[tb]
\centering
\begin{tabular}{l | c | c}
\toprule
\textbf{Test} & \textbf{BERT with Vision} & \textbf{EXPERT}  \\ \midrule
\multicolumn{3}{c}{Acquiring Familiar Words} \\ \midrule
Seen verbs & 68.5 & 71.7 \\
Seen nouns & 44.6 & 59.3 \\
Seen compositions & 35.7 & 40.9 \\
New compositions & 30.0 & 34.4 \\
\midrule
\multicolumn{3}{c}{Acquiring New Words} \\ \midrule
Nouns 1:1 & 56.2 & 78.6 \\
Verbs 1:1 & 60.5 & 72.1 \\
Nouns 2:1 & 50.7 & 71.5 \\
Verbs 2:1 & 44.1 & 64.5 \\
\bottomrule
\end{tabular}
\vspace{-0.5em}
\caption{\textbf{Results on EPIC-Kitchens without Bounding Boxes:} We show accuracy on all evaluation metrics used in the paper, but withhold ground truth object bounding boxes at test time. EXPERT continues to outperform the competitive vision and language baseline.\vspace{-1em}}
\label{tbl:nobboxes_app} 
\end{table}

\subsection{Does EXPERT require bounding boxes?} 
\label{bboxes}
We test EXPERT on EPIC-Kitchens without any additional image regions (i.e., only with the full image) to quantify the importance of providing grouped image regions. This model experiences a slight performance decrease in language acquisition, performing 4\% worse with a 1:1 distractor ratio, and 2\% worse with a 2:1 ratio. However, the baseline BERT with Vision model has larger decreases on all metrics when tested without additional image regions, so our model still outperforms it. When testing on masked language modeling (as in Sections 4.4 and 4.5 in the main paper), EXPERT performs 10\% worse on verbs, 14\% worse on nouns, 24\% on seen compositions, and 29\% on new compositions. However, the baseline model again has larger decreases in performance. These experiments show that while EXPERT improves when provided more visual information, it can acquire new language even when provided just with one image and perform stronger than baselines. 

On Flickr30k, when testing with only the full image as model input in addition to text, performance decreases by \emph{at most 7\%}. Therefore, EXPERT does not require using image regions at test time. Performance decreases by a similar amount on vision and language baselines, such that EXPERT still outperforms them.

\clearpage
\subsection{Flickr Visualizations}
\label{flickr_viz}
\begin{figure*}[!hb]
\includegraphics[width=\linewidth]{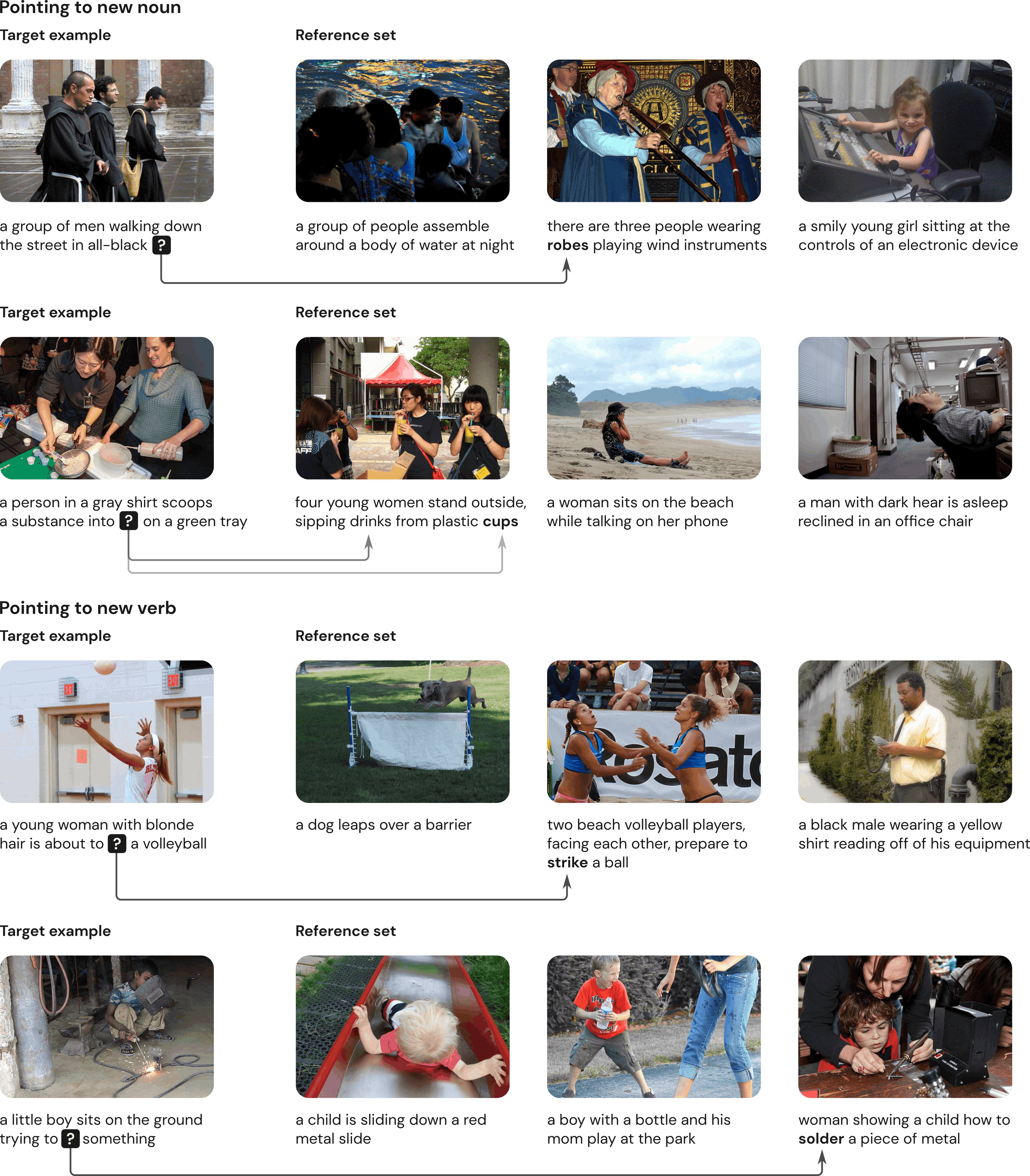}
\caption{\textbf{Acquiring New Words on Flickr30k:}  We show examples where the model acquires new words.
{\setlength{\fboxsep}{2pt}\colorbox{black}{\textcolor{white}{?}}} in the target example indicates the masked out new word. \textbf{Bold} words in the reference set are ground truth. The model makes predictions by pointing into the reference set, and the weight of each pointer is visualized by the shade of the arrows shown (weight $< 3\%$ is omitted). }
\label{fig:morelm_app} 
\end{figure*}

\clearpage

\begin{figure*}[!hb]
\includegraphics[width=\linewidth]{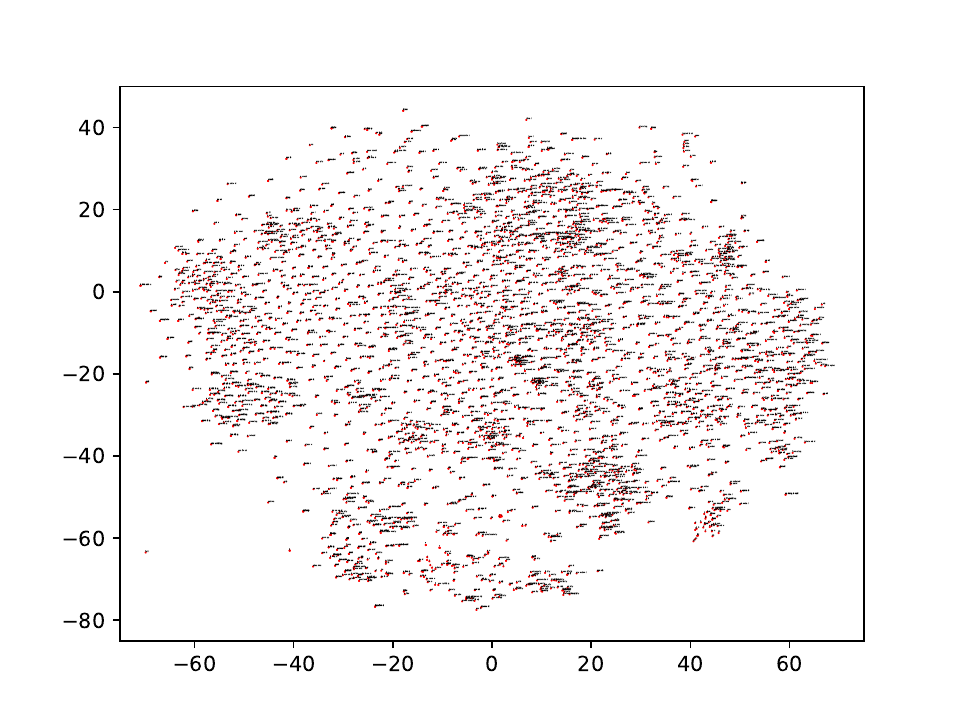}
\caption{\textbf{t-SNE 2D projection of Flickr embeddings:} We show a 2D projection computed using the t-SNE algorithm \cite{tsne} of the word embedding matrix for EXPERT trained on the Flickr dataset. Each dot corresponds to the word shown to its top right. Please zoom in to view in detail.}
\label{fig:tsne_app} 
\end{figure*}

\end{document}